\documentclass[runningheads]{llncs}
\usepackage{latexsym}

\usepackage{times}
\usepackage{soul}
\usepackage{url}
\usepackage[hidelinks]{hyperref}
\usepackage[utf8]{inputenc}
\usepackage{graphicx}
\usepackage{amsmath}
\usepackage{booktabs}
\usepackage{algorithm}
\usepackage{algorithmic}
\usepackage{url,ifthen}
\usepackage{amssymb}

\usepackage{graphicx}
\usepackage[export]{adjustbox}
\usepackage{subcaption}
\usepackage[section]{placeins}
\usepackage[T1]{fontenc}

\begin{document}

\title{Monte Carlo Inverse Folding}

\author{Tristan Cazenave \and Thomas Fournier}

\authorrunning{T. Cazenave and T. Fournier}

\institute{LAMSADE, Université Paris-Dauphine, PSL, CNRS, France \email{Tristan.Cazenave@dauphine.psl.eu}}

\maketitle

\begin{abstract}
The RNA Inverse Folding problem comes from computational biology. The goal is to find a molecule that has a given folding. It is important for scientific fields such as bioengineering, pharmaceutical research, biochemistry, synthetic biology and RNA nanostructures. Nested Monte Carlo Search has given excellent results for this problem. We propose to adapt and evaluate different Monte Carlo Search algorithms for the RNA Inverse Folding problem.
\end{abstract}

\section{Introduction}

Monte Carlo Tree Search (MCTS) \cite{Coulom2006,Kocsis2006} originated in the field of computer Go and has been applied to many other problems \cite{BrownePWLCRTPSC2012}.

Nested Monte Carlo Search (NMCS) \cite{CazenaveIJCAI09} uses multiple levels of search, memorizing the best sequence of each level. It has been applied to many single-player games and optimization problems \cite{RimmelEvo11,Mehat2010,kinny2012new,Bouzy13,Cazenave13Discovery,PouldingF14,PouldingF15,Bouzy16} and also to two-player games \cite{cazenave2016nested}.

Nested Rollout Policy Adaptation (NRPA) \cite{Rosin2011} also uses multiple levels of search, memorizing the best sequences. It additionally learns a playout policy using the best sequences. It has also been applied to many problems \cite{cazenave2012tsptw,edelkamp2013algorithm,edelkamp2014monte,edelkamp2014solving,edelkamp2015monte,edelkamp2016monte,Edelkamp16Diversity} and games \cite{cazenave2016tcs}.

The RNA design problem also named the RNA Inverse Folding problem is computationally hard \cite{Bonnet2020Designing}. This problem is important for scientific fields such as bioengineering, pharmaceutical research, biochemistry, synthetic biology and RNA nanostructures \cite{portela2018unexpectedly}. NMCS has been successfully applied to the RNA Inverse Folding problem with the NEMO program \cite{portela2018unexpectedly}. As a follow-up to NEMO, we propose to investigate different Monte Carlo Search algorithms for this problem.

The paper is organized as follows. The second section describes the Inverse Folding problem, the NEMO program by Fernando Portela \cite{portela2018unexpectedly} and the domain knowledge used in NEMO. The third section describes the Monte Carlo Search algorithms we have used for solving Inverse Folding problems of the Eterna100 benchmark. We present a new algorithm performing well for this problem, the GNRPA algorithm with restarts. The fourth sections details experimental results.

\section{Inverse Folding}

\subsection{Presentation of the RNA inverse folding problem}

An RNA strand is a molecule composed of a sequence of nucleotides. This strand folds back on itself to form what is called its secondary structure (See Figure 1). It is possible to find in a polynomial time the folded structure of a given sequence. However, the opposite, which is the RNA inverse folding problem, is much harder and is supposed to be NP-complete. This problem still resists algorithmic approaches that still fail to match the performance of human experts. This is partly due to the chaotic changes that can occur in the secondary structure because of a small change in the sequence (See difference between Figure 3 and 2.b).

These performances are evaluated on the Eterna100 benchmark which contains 100 RNA secondary structure puzzles of varying degrees of difficulty. A puzzle consists of a given structure under the dot-bracket notation. This notation defines a structure as a sequence of parentheses and points each representing a base. The matching parentheses symbolize the paired bases and the dots the unpaired ones. The puzzle is solved when a sequence of the four nucleotides A,U,G and C, folding according to the target structure, is found. In some puzzles, the value of certain bases is imposed.

Where human experts have managed to solve the 100 problems of the benchmark, no program has so far achieved such a score. The best score so far is 95/100 by NEMO, NEsted MOnte Carlo RNA Puzzle Solver \cite{portela2018unexpectedly}.

\begin{figure}[htp]
    \centering
    \includegraphics[width=10cm]{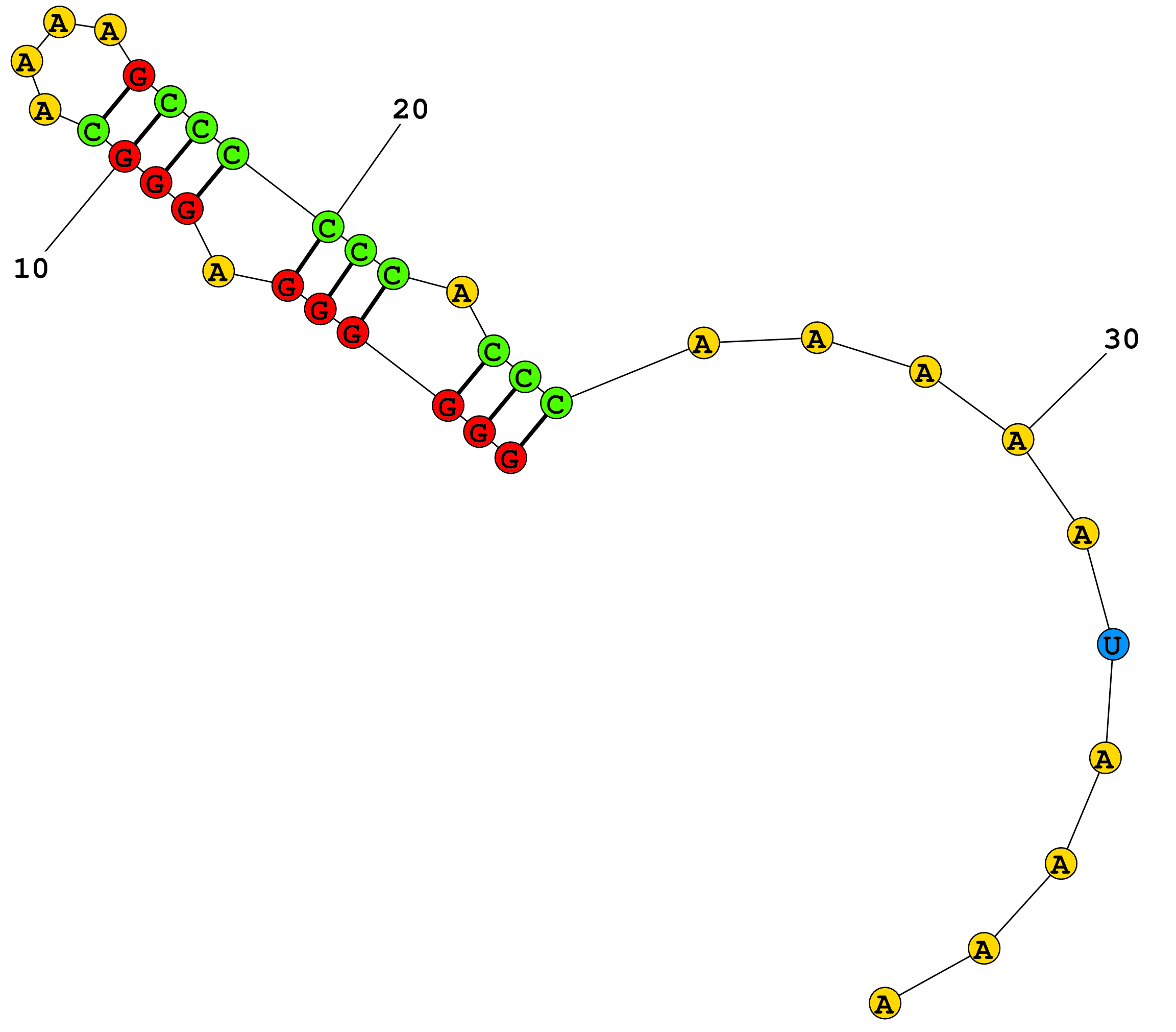}
    \caption{Secondary Structure of a RNA strand}
    \label{fig:figure1}
\end{figure}

\begin{figure}
\begin{subfigure}{0.5\textwidth}
\includegraphics[width=0.9\linewidth, height=8cm]{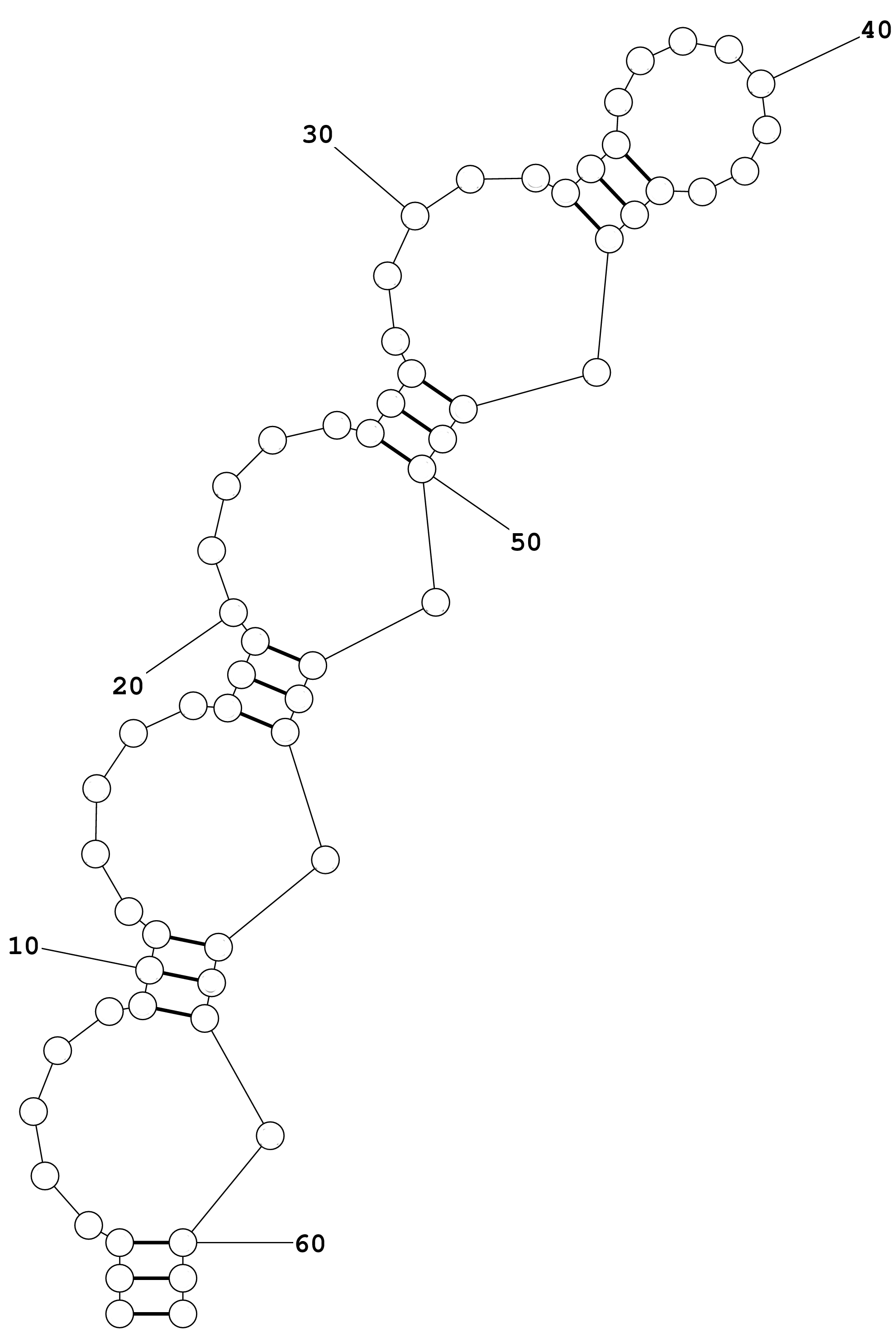} 
\caption{Target Structure of a Puzzle}
\end{subfigure}
\begin{subfigure}{0.5\textwidth}
\includegraphics[width=0.9\linewidth, height=8cm]{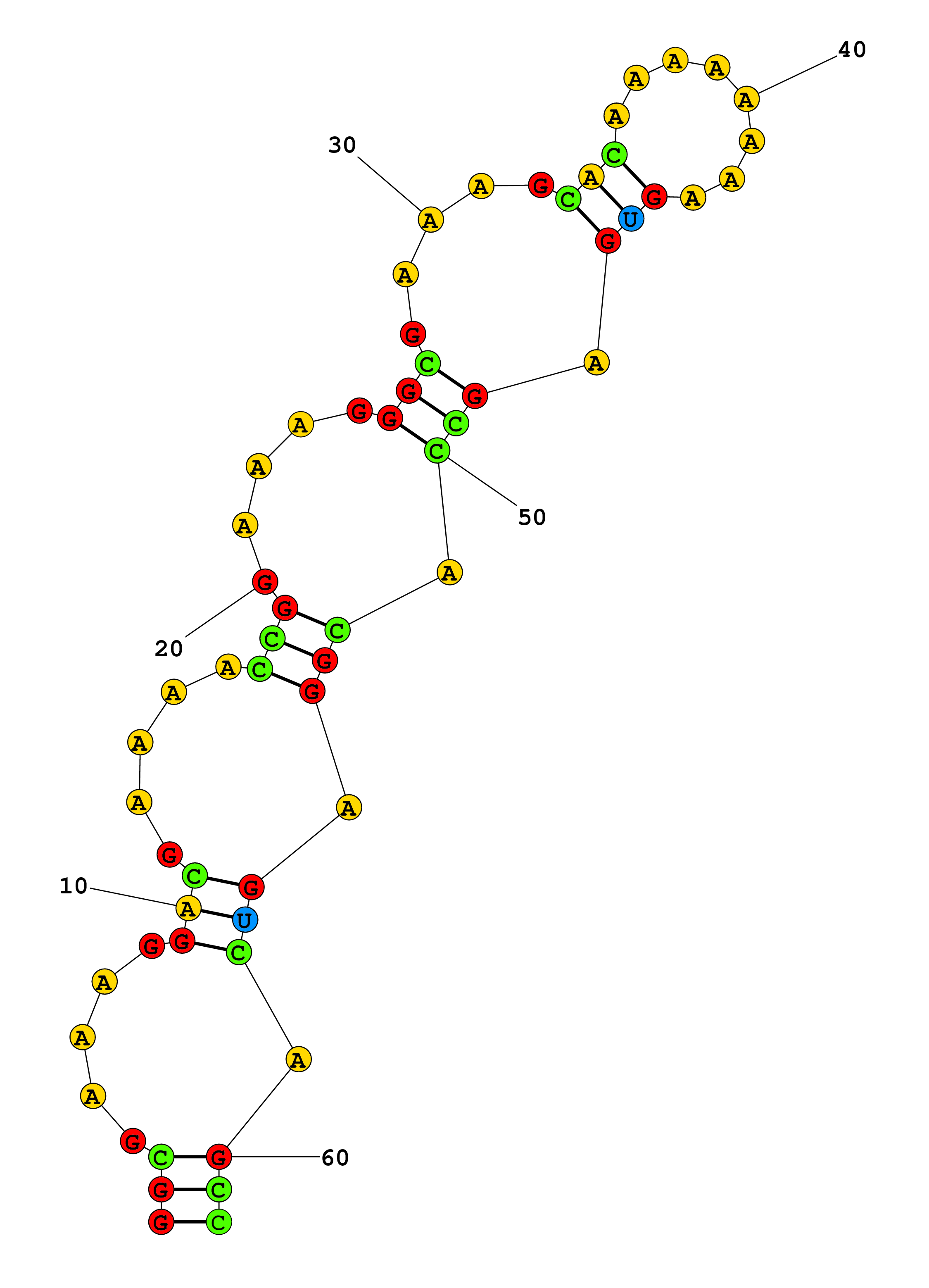}
\caption{Solved puzzle}
\end{subfigure}
\caption{Exemple of an eterna100 puzzle}
\label{fig:image2}
\end{figure}

\begin{figure}
    \centering
    \includegraphics[width=6cm]{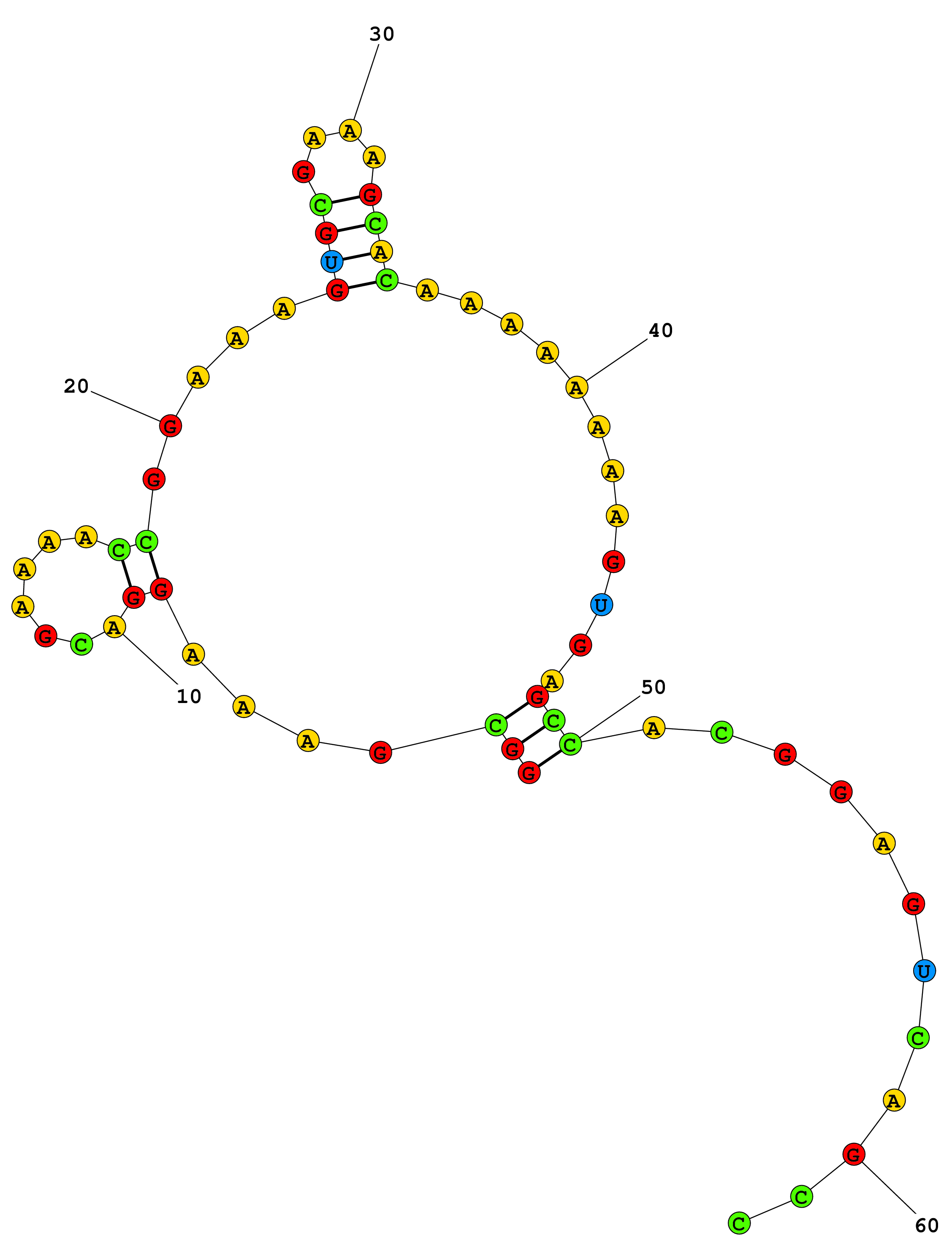}
    \caption{Sequence of figure 2.b with base 25 replaced by U}
\end{figure}

\subsection{NEMO}

NEMO works by performing several iterations of NMCS-B, a slightly modified version of NMCS. Between each iteration of the NMCS-B, NEMO retains part of the best current solution. It identifies a part of the sequence on which to perform mutations using stochastic heuristics and restarts the NMCS-B on it. We model a candidate by a sequence. The bases assigned are represented by the corresponding letter (A/U/G/C) and the others by the letter N. This is how the NMCS-B identifies the bases on which it must work.

NEMO uses a level 1 NMCS for its NMCS-B. At level 1, the NMCS, in each state of the problem, will perform a certain number of playouts for each possible move. It then plays the move that led to the best playout and moves to the next state until it reaches a final state. In the context of NEMO, each state corresponds to a sequence, initially the candidate sequence. Each move consists in taking the first N in the sequence and assigning a value to it, working on paired bases first. When the base is paired in the target structure, it will assign a value to both bases of the pair simultaneously. Indeed only the three combinations AU, GC, and GU can be paired, so it is more convenient to consider them at the same time. To perform playouts, NEMO is also using heuristics with biased weights depending on the location in the target structure. In addition, unlike the classic NMCS, the NMCS-B retains the best playout achieved so far throughout the execution. A final state is found when the sequence is fully completed. The playouts are evaluated according to the function :

$$
\begin{array}{cc}
     &  
score = \left\{
    \begin{array}{ll}
        \frac{K}{1+\Delta G} & \mbox{if } K>0 \\
        K(1+\Delta G) & \mbox{else}
    \end{array}
\right. \\
& \\
&
    \mbox{with }
    K=1-\frac{BPD}{2*NumTargetPairs}
\end{array}
$$

Where $BPD$ is the number of different pairs between the secondary structure of the sequence and the target structure. \\ 
$NumTargetPairs$ is the number of pairs in the target structure.\\
$\Delta G$ is the difference between the Minimum Free Energy of the secondary structure and the free energy that the sequence would have in the target structure.

\subsection{Domain Knowledge and Heuristics}
The heuristics used for the sampling of the NMCS-B are based on domain knowledge and personal experience and are chosen without computational optimization. \\
The paired bases are generally chosen according to the same rule. With the exception of adjacent stacks in multi-loops, the closing pairs of the left-most and rightmost stacks are chosen with the weights given in the following table.
The notion of right and left in this case is defined from the point of view of the inside of the loop.
\begin{center}
 \centering
 \begin{tabular}{||c | c c c||} 
 \hline
 Paired Bases & GC/CG & AU/UA & GU/UG \\ [0.5ex] 
 \hline\hline
 General Case & 60\% & 33\% & 7\% \\ 
 \hline
 Left-Most in Junction & 82\% & 11\% & 7\% \\
 \hline
 Right-Most in Junction & 37\% & 56\% & 7\% \\
 \hline
\end{tabular}
\end{center}

Various rules are applied for unpaired bases. The weights used to choose between A/U/G/C in the general case are 93\%, 1\%, 5\% and 1\%. Mismatches are treated differently depending on the case.\\
Since NEMO first assigns a value to the paired bases, the weights for the bases with a paired mismatch depend on the value of this mismatch.
\begin{center}
 \centering
 \begin{tabular}{||c | c c c c ||} 
 \hline
 Mismatch with a Paired Base & A & U & G & C \\ [0.5ex] 
 \hline\hline
 Mismatch is a paired A & 63\% & 0\% & 25\% & 12\%\\ 
 \hline
 Mismatch is a paired U & 0\% & 55\% & 9\% & 36\% \\ 
 \hline
 Mismatch is a paired G & 25\% & 12\% & 63\% & 0\% \\
 \hline
 Mismatch is a paired C & 55\% & 36\% & 0\% & 9\% \\
 \hline
\end{tabular}
\end{center}

Furthermore, in internal loops, the weights also depend on the mismatch value if it has already been assigned, otherwise a more general rule applies.

\begin{center}
 \centering
 \begin{tabular}{||c | c c c c ||} 
 \hline
 Mismatch in Internal Loops & A & U & G & C \\ [0.5ex] 
 \hline\hline
 Mismatch is not assigned & 18\% & 4\% & 74\% & 4\% \\ 
 \hline
 Mismatch is A & 44\% & 0\% & 44\% & 12\%\\ 
 \hline
 Mismatch is U & 0\% & 67\% & 11\% & 22\% \\ 
 \hline
 Mismatch is G & 67\% & 11\% & 22\% & 0\% \\
 \hline
 Mismatch is C & 66\% & 17\% & 0\% & 17\% \\
 \hline
\end{tabular}
\end{center}

Finally, the mismatch in junctions and external loops are drawn according to the distribution 97\%, 1\%, 1\% and 1\%.\\
Much stronger and more deterministic rules are applied with high probabilities (more than 80\%) in specific cases, especially for triloops and internal loops. For both 1-1 and 2-2 internal loops for instance, only one mismatch pair is possible, and there are only three possibilities in the general case. This is part of a process of reproducing a "boosting" strategy. Depending on the type of loop, certain combinations of nucleotides at specific locations called "boosting points", especially terminal mismatches, can be used to reduce the energy of the structure. However, the most difficult puzzles may require less conventional solutions, hence the need not to apply these rules 100\% of the time.\\
Therefore, in the use we will make of this heuristic we will not apply these last rules and when we mention the weights of the NEMO heuristic we refer to the previously mentioned values. \\
\\
In addition, between iterations of NMCS-B, if it hasn't solved the problem NEMO keeps part of the best current solution to restart the algorithm on. The set of bases that are not kept contains those that do not fold correctly, their neighborhood and randomly selected bases. This principle has not been applied to the presented algorithms.

\section{Monte Carlo Search}

\subsection{Presentation of the NRPA algorithm}

    The Nested Rollout Policy Adaptation (NRPA) algorithm is a Monte Carlo Tree Search based algorithm with adaptive rollout policy during execution. It is a recursive algorithm. At level 0 it generates a playout according to the current policy. At level n, it calls for a given number of iterations the n-1 level of the algorithm, adapting the policy each time with the best solution so far. NRPA is given in algorithms \ref{PLAYOUT}, \ref{ADAPT} and \ref{NRPA}.

\begin{algorithm}
\begin{algorithmic}[1]
\STATE{playout ($state$, $policy$)}
\begin{ALC@g}
\STATE{$sequence$ $\leftarrow$ []}
\WHILE{true}
\IF{$state$ is terminal}
\RETURN{(score ($state$), $sequence$)}
\ENDIF
\STATE{$z$ $\leftarrow$ 0.0}
\FOR{$m$ in possible moves for $state$}
\STATE{$z$ $\leftarrow$ $z$ + exp ($policy$ [code($m$)])}
\ENDFOR
\STATE{choose a $move$ with probability $\frac{exp (policy [code(move)])}{z}$}
\STATE{$state$ $\leftarrow$ play ($state$, $move$)}
\STATE{$sequence$ $\leftarrow$ $sequence$ + $move$}
\ENDWHILE
\end{ALC@g}
\end{algorithmic}
\caption{\label{PLAYOUT}The playout algorithm}
\end{algorithm}

\begin{algorithm}
\begin{algorithmic}[1]
\STATE{Adapt ($policy$, $sequence$)}
\begin{ALC@g}
\STATE{$polp \leftarrow$ $policy$}
\STATE{$state \leftarrow$ $root$}
\FOR{$move$ in $sequence$}
\STATE{$polp$ [code($move$)] $\leftarrow$ $polp$ [code($move$)] + $\alpha$}
\STATE{$z$ $\leftarrow$ 0.0}
\FOR{$m$ in possible moves for $state$}
\STATE{$z$ $\leftarrow$ $z$ + exp ($policy$ [code($m$)])}
\ENDFOR
\FOR{$m$ in possible moves for $state$}
\STATE{$polp$ [code($m$)] $\leftarrow$ $polp$ [code($m$)] - $\alpha * \frac{exp (policy [code(m)])}{z}$}
\ENDFOR
\STATE{$state$ $\leftarrow$ play ($state$, $move$)}
\ENDFOR
\STATE{$policy$ $\leftarrow$ $polp$}
\end{ALC@g}
\end{algorithmic}
\caption{\label{ADAPT}The Adapt algorithm}
\end{algorithm}

\begin{algorithm}
\begin{algorithmic}[1]
\STATE{NRPA ($level$, $policy$)}
\begin{ALC@g}
\IF{level == 0}
\RETURN{playout (root, $policy$)}
\ELSE
\STATE{$bestScore$ $\leftarrow$ $-\infty$}
\FOR{N iterations}
\STATE{(result,new) $\leftarrow$ NRPA($level-1$, $policy$)}
\IF{result $\geq$ bestScore}
\STATE{bestScore $\leftarrow$ result}
\STATE{seq $\leftarrow$ new}
\ENDIF
\STATE{policy $\leftarrow$ Adapt (policy, seq)}
\ENDFOR
\RETURN{(bestScore, seq)}
\ENDIF
\end{ALC@g}
\end{algorithmic}
\caption{\label{NRPA}The NRPA algorithm.}
\end{algorithm}

\subsection{Application of NRPA to Inverse Folding}

As part of the Inverse Folding problem, one solution consists in a chain of bases. A playout is made by running through the targeted structure of the chain, each move consists in assigning a value to the missing links. We distinguish between two cases, the unpaired bases in the target structure have four possible moves, one for each nucleic base (A/U/G/C). The unpaired bases have 6 possible moves, one for each possible ordered combination with their pair (GC/CG/AU/...). Each move is therefore defined by its position in the chain and whether it is a pair or not. Thus, there is a fixed number of moves, which are always ordered in the same way. Solutions are evaluated with the same score function as in the NEMO algorithm which is a combination of the fitness of the chain with the target structure and the difference between the target structure and the folded chain structure.

\subsection{GNRPA}

Let $w_{ib}$ be the weight associated to move b at index i in the sequence.
In NRPA the probability of choosing move b at index i is:

$$p_{ib} = \frac{e^{w_{ib}}}{\Sigma_k e^{w_{ik}}}$$

We propose to try GNRPA \cite{Cazenave2020GNRPA} for Inverse Folding and to replace it with:

$$p_{ib} = \frac{e^{w_{ib}+\beta_{ib}}}{\Sigma_k e^{w_{ik}+\beta_{ik}}}$$

where we use for $\beta_{ib}$ the logarithm of the  probabilities used in Nemo.

\subsection{Stabilized GNRPA}

Stabilized NRPA \cite{Cazenave2020Stabilized} is a simple improvement of NRPA. The principle is to play P playouts at level 1 before each call to the adapt function. The number of calls to the adapt function as level 1 is still N, the number of iteration of upper levels. So at level 1, $P \times N$ playouts are performed.

\subsection{Beam GNRPA}

Beam NRPA has already been applied successfully to the TSPTW and to Morpion Solitaire \cite{Cazenave2012BeamNRPA}. The best results were obtained using a beam at level 1. Similarly to Stabilized GNRPA, at level 1, $B \times N$ playouts are performed for a beam of size B. However the algorithm is different from Stabilized NRPA since it memorizes B best sequences and B policies and plays the B playouts with different policies.

As Stabilized GNRPA it is embarrassingly parallel at level 1 and can be very efficient on a parallel machine.

When using Beam NRPA it can be beneficial to ensure the diversity of the beam \cite{Edelkamp16Diversity}. The diversity criterion we have used is to only keep in the beam sequences that have different scores. It is simple and efficient as it ensures diversity while keeping enough sequences.

\subsection{Coding Moves}

The natural way to code moves for Inverse Folding is to use the index of the base or of the pair of base in the string (m.index) and the index of the base in the list of bases or of the pair of bases in the list of pairs of bases (m.number). The formula is then:

$$code (m) = m.index + M \times m.number$$

For example if the move $m$ is to put the fourth base at index 10 the code is $10 + 2000 *4 = 8010$, provided strings always have less than 2000 characters and therefore $M=2000$.

It may be interesting to include the previously chosen bases in the code of a move, for example if a base has meaning only if following another base. We call the history of a code the number of bases in the history included in the code. The previous formula holds for a code history of 0. The code for a code history of 1 is:

$$code (m) = m.index + M \times m.number + M \times 6 \times previousMove.number$$

Six is the maximum value for a move number, the maximum number of legal moves is 6. The code for a code history of 2 includes the two previous moves in the code.

\subsection{Start Learning}

In order to wait for better sequences before learning it is possible to delay learning only after a given number of sequences have been found \cite{cazenave2016selective}.

\subsection{Zobrist Hashing}

Each state is associated to a different hash code. This is done with Zobrist Hashing. Each move at each index is associated to a random number. The hash code of a state is the XOR of all the random numbers corresponding to the moves that have been played to reach this state. Zobrist hashing is used in games to build a transposition table. We will use it in the UCT variants. Another use of Zobrist Hashing is to detect playouts that have already been evaluated. As most of the time in Inverse Folding is spent scoring the playouts, it is advisable to avoid reevaluating an already evaluated playouts. This is done with a score hash table that contains the hash of the terminal states already encountered associated to their computed scores. If a terminal state is met again the score need not be recomputed it can just be sent back from the table. There are variations on the number of playouts already evaluated according to the sequence, some sequences have very few while others have a lot.

\subsection{Restarts}

There can be large variations on the solving times of some problems. Sometimes the search algorithm takes a wrong direction and stay stuck on a suboptimal sequence without making any progress. A way to deal with this behavior is to periodically stop and restart the search. For some difficult problems however a long search is required to find the solution. There are multiple ways to use restarts. The algorithm can double the search time at each restart for example. We call this method iterative doubling. We have observed that a level 2 search is able to solve many problems, a restart strategy can also be to repeatedly call GNRPA at level 2 until thinking time is elapsed. Another way to deal with search being stuck is to stop a level when the best sequence has not changed for a given number of recursive calls.

It is difficult to set a static restart strategy for all problems. Long sequences are much more difficult than short ones and the progress on long sequences is slower. In order to cope with this property we use a restart threshold. It is set to the length of the sequence divided by 5. When using this restart strategy there is no limit on the length of a level.

Algorithm \ref{restarts} gives the GNRPA algorithm with restarts.


\begin{algorithm}
\begin{algorithmic}[1]
\STATE{GNRPA ($level$, $policy$)}
\begin{ALC@g}
\IF{level == 0}
\RETURN{playout (root, $policy$)}
\ELSE
\STATE{$bestScore$ $\leftarrow$ $-\infty$}
\STATE{$last \leftarrow 0$}
\FOR{i in range($\infty$)}
\STATE{(result,new) $\leftarrow$ GNRPA($level-1$, $policy$)}
\IF{result $\geq$ bestScore}
\IF{result $>$ bestScore}
\STATE{$last \leftarrow i$}
\ENDIF
\STATE{bestScore $\leftarrow$ result}
\STATE{seq $\leftarrow$ new}
\ENDIF
\IF{$i - last \geq limitRestart$}
\RETURN{(bestScore, seq)}
\ENDIF
\IF{$i \geq startLearning$}
\STATE{policy $\leftarrow$ Adapt (policy, seq)}
\ENDIF
\ENDFOR
\ENDIF
\end{ALC@g}
\end{algorithmic}
\caption{\label{restarts}The GNRPA algorithm with restarts.}
\end{algorithm}

\subsection{Parallelization}

The multiple playouts of stabilized NRPA and the loop over the elements of the beam are embarrassingly parallel. We simply parallelized with OpenMP a common loop including the beam and the stabilized playouts at level 1. If we have 4 stabilized playouts and a beam of 8, the 32 resulting playouts are played in parallel. This kind of parallelization is a kind of leaf parallelization \cite{cazenave2007CGW,chaslot2008parallel}.

We also experiment with root parallelization \cite{cazenave2007CGW,chaslot2008parallel}. The principle is to perform multiple independent GNRPA in parallel and to stop as soon as one has found a solution or when the allocated time is elapsed.

Leaf parallelization is more difficult to scale than root parallelization. For the same wall clock time leaf parallelization runs more iterations for a single policy than root parallelization which optimizes many more different policies but with less iterations. On the other hand root parallelization scales very well and has a built-in restart strategy. Root parallelization works well for problems that converge relatively rapidly on a suboptimal solution as they benefit from restarts, while leaf parallelization works better for problems that converge slowly but steadily towards the best solutions.










\begin{table}
  \centering
  \caption{Number of problems solved, out of 100 problems, using different parameters.}
  \label{parameters}
  \begin{tabular}{rrrrrrrrrrr}
  Level & $\alpha$ & N & $\beta_{ib}$ & P & ~~~~Beam & ~~~H & ~~~~~Solved \\
~~~~~~~~~~&  ~~~~~~~~~~~~&  ~~~~~~~~~~~~ & ~~~~~~~~~~ &~~~~~~~~~ & ~~~~~~~~~ &  & &~~~~~~~~~~ \\
        1 &          1.0 &          100 &           no &         1 & 1.1.1 & 0 &  3 \\
        1 &          1.0 &          100 &          yes &         1 & 1.1.1 & 0 & 30 \\
        1 &          1.0 &          100 &          yes &         1 & 1.1.1 & 1 & 32 \\
        1 &          1.0 &          100 &          yes &         1 & 4.1.1 & 1 & 42 \\
        1 &          1.0 &          100 &          yes &         1 & 8.1.1 & 1 & 53 \\
        1 &          1.0 &          100 &          yes &         1 & 16.1.1 & 1 & 54 \\
        1 &          1.0 &          100 &          yes &         4 & 8.1.1 & 1 & 69 \\
        1 &          1.0 &          100 &          yes &         4 & 8.1.1 & 2 & 69 \\
        2 &          1.0 &          100 &           no &         1 & 1.1.1 & 0 & 49 \\
        2 &          1.0 &          100 &          yes &         1 & 1.1.1 & 0 & 73 \\
        2 &          1.0 &          100 &          yes &         1 & 1.1.1 & 1 & 75 \\
        2 &          1.0 &          100 &          yes &         2 & 1.1.1 & 0 & 73 \\
        2 &          1.0 &          100 &          yes &         3 & 1.1.1 & 0 & 74 \\
        2 &          1.0 &          100 &          yes &         4 & 1.1.1 & 0 & 80 \\
        2 &          1.0 &          100 &          yes &         5 & 1.1.1 & 0 & 77 \\
        2 &          1.0 &          100 &          yes &         6 & 1.1.1 & 0 & 75 \\
        2 &          1.0 &          100 &          yes &         7 & 1.1.1 & 0 & 80 \\
        2 &          1.0 &          100 &          yes &         8 & 1.1.1 & 0 & 79 \\
        2 &          1.0 &          100 &          yes &         9 & 1.1.1 & 0 & 81 \\
        2 &          1.0 &          100 &          yes &        10 & 1.1.1 & 0 & 80 \\
        2 &          1.0 &          100 &          yes &         4 & 8.1.1 & 1 & 85 \\
        3 &          1.0 &          100 &          yes &         1 & 1.1.1 & 0 & 85 \\
  \end{tabular}
\end{table}

\begin{table}
  \centering
  \caption{Parallelization efficiency.}
  \label{parallel}
  \begin{tabular}{lrrrrrrrrrr}
 Algorithm & 1 & 2 & 4 & 6 & 8 & 12 \\
 ~~~~~~~~~~&  ~~~~~~~~~~~~&  ~~~~~~~~~~~~&   ~~~~~~~~~~ &~~~~~~~~~~ & ~~~~~~~~~~ & ~~~~~~~~~~ & ~~~~~~~~~~ & ~~~~~~~~~~ & ~~~~~~~~~~ & ~~~~~~~~~~ \\
GNRPA(level=1,N=100,P=4,Beam=8) & 11.916 & 6.889 & 4.526 & 3.657 & 3.169 & 3.359
  \end{tabular}
\end{table}

\begin{table}
  \centering
  \caption{Number of problems solved by GNRPA using different parameters and a fixed time limit.}
  \label{time}
  \begin{tabular}{lrrrrrrrrrrrrrrr}
 $\beta_{ib}$ & P &  Beam & ~~// & ~~~R & ~~~~~~~~~~~~N & ~~~Start & ~~~H & ~1m & ~2m & ~4m & ~8m & 16m & 32m & 64m \\
  \\
           no & 1.1 & 1.1 & n & n & 100.100 & 0.0 &       0 & 30 & 33 & 41 & 50 & 61 & 67 & 69 \\
          yes & 1.1 & 1.1 & n & n & 100.100 & 0.0 &       0 & 58 & 64 & 68 & 72 & 74 & 75 & 79 \\
          yes & 4.1 & 4.1 & n & n & 100.100 & 0.0 &       0 & 71 & 75 & 78 & 79 & 81 & 83 & 84 \\
          yes & 4.1 & 8.1 & n & n & 100.100 & 0.0 &       0 & 75 & 75 & 79 & 80 & 81 & 83 & 84 \\
          yes & 4.1 & 8.1 & n & n & 100.100 & 0.0 &       1 & 75 & 78 & 80 & 82 & 83 & 85 & 87 \\
          yes & 4.1 & 8.1 & n & n & 100.100 & 4.4 &       1 & 76 & 80 & 81 & 82 & 84 & 85 & 87 \\
          yes & 4.1 & 8.1 & y & n & 100.100 & 4.4 &       1 & 78 & 84 & 83 & 86 & 87 & 87 & 88 \\
          yes & 4.1 & 8.1 & y & 3 & $\infty.\infty$ & 4.4 &       1 & 80 & 84 & 85 & 85 & 88 & 89 & 92 \\
 \end{tabular}
\end{table}

\begin{table}
  \centering
  \caption{Number of problems solved by different algorithms and a fixed time limit.}
  \label{other}
  \begin{tabular}{lrrrrrrrrrr}
 Algorithm & 1m & 2m & 4m & 8m & 16m & 32m & 64m \\
 ~~~~~~~~~~&  ~~~~~~~~~~~~&  ~~~~~~~~~~~~&   ~~~~~~~~~~ &~~~~~~~~~~ & ~~~~~~~~~~ & ~~~~~~~~~~ & ~~~~~~~~~~  \\
UCT(0.4)                & 47 & 49 & 50 & 54 & 56 & 57 & 62 \\
Nested(1)               & 59 & 62 & 64 & 65 & 67 & 71 & 73 \\
Nested(2)               & 55 & 59 & 61 & 66 & 66 & 71 & 73 \\
Diversity GNRPA(5.1)    & 72 & 73 & 73 & 77 & 78 & 78 & 80 \\
 \end{tabular}
\end{table}


\begin{table}[ht]
  \centering
  \caption{Number of problems solved with different options.}
  \label{order}
  \begin{tabular}{rrrrrrrrrrr}
 Correction & Order & 1m & 2m & 4m & 8m & 16m \\ 
 ~~~~~~~~~~&  ~~~~~~~~~~~~&  ~~~~~~~~~~~~&   ~~~~~~~~~~ &~~~~~~~~~~ & ~~~~~~~~~~ & ~~~~~~~~~~ \\
 no &  no & 84 & 87 & 85 & 85 & 87 &  &  \\
yes &  no & 80 & 86 & 85 & 86 & 87 &  &  \\
 no & yes & 81 & 84 & 86 & 85 & 87 &  &  \\
yes & yes & 80 & 82 & 84 & 88 & 87 &  &  \\
 \end{tabular}
\end{table}

\begin{table}[ht]
  \centering
  \caption{Number of problems solved with root parallel GNRPA.}
  \label{root}
  \begin{tabular}{rrrrrrrrrrr}
 Process & 1m & 2m & 4m & 8m & 16m & 32m & 64m \\
 ~~~~~~~~~~&  ~~~~~~~~~~~~&  ~~~~~~~~~~~~&   ~~~~~~~~~~ &~~~~~~~~~~ & ~~~~~~~~~~ & ~~~~~~~~~~ & ~~~~~~~~~~ & ~~~~~~~~~~  \\
20 & 82 & 84 & 85 & 86 & 87 & 88 & 89 \\
 \end{tabular}
\end{table}

\section{Experimental Results}

We now detail experiments with the different Monte Carlo Search algorithms on the Eterna100 benchmark.







Table \ref{parameters} gives the number of problems solved with different parameters for the GNRPA algorithm. We can see that at level 1 using GNRPA instead of NRPA enables to solve 30 problems instead of 3. Similarly at level 2 it solves 73 problems instead of 49. Using Stabilized GNRPA with P=4 and Beam GNRPA with a beam of 8 at level 1 also improves quite much the number of problems solved at levels 1 and 2. Interestingly for level 2 it is 32 times slower that a regular level 2 search and solves 85 problems, when a search at level 3 is 100 times slower and still solves 85 problems.

Table \ref{parallel} gives the median time over 3 runs of a level 1 search with different numbers of threads on problem 64 which is difficult. The algorithm is GNRPA with P = 4, a beam of 8 and N = 100. We can see that using 8 threads gives the best results. We optimized memory in order to avoid cache misses but we were not able to have better results with more threads.

Table \ref{time} gives the number of problems solved within a fixed time limit for different algorithms. The time limits range from 1 minute to 64 minutes per problem. The parallel algorithm runs on a multicore machine. The results of the parallel program are given in the last line of table \ref{time}. The times used to stop the parallel program are the wall clock times. The last line is leaf parallel GNRPA with restarts and gives the best results within 1 hour of wall clock time, solving 92 of the 100 problems in one run. Start is 4 meaning that it starts learning after 4 playouts. H is 1 meaning the code include the previous move, R is 3 meaning that the restart threshold is set to the length of the string divided by 3.


The problems solved by different runs of 1 hour we made are not always the same. Some hard problems are solved only in some runs. So the limit of 92 solved problems is not the limit of the algorithm. The problems that were never solved in 1 hour are problems 100, 99, 97, 91, 90, 78. With a two hours limit, problem 90 is solved thus reaching 95 solved problems, the same number of solved problems as NEMO.

For the sake of completeness we also tested other popular Monte Carlo Search algorithms. The results are given in table \ref{other} for UCT \cite{Kocsis2006}, Nested Monte Carlo Search \cite{CazenaveIJCAI09,portela2018unexpectedly} and Diversity NRPA \cite{Edelkamp16Diversity}. The UCT constant is set to 0.4, NMCS is tested for repeated calls to level 1 and level 2. Diversity GNRPA \cite{Edelkamp16Diversity} is called with a set of 5 sequences at level 1 and 1 sequence at level 2. The improved GNRPA algorithm gives better results than these algorithms.

Table \ref{order} gives the results with time of the best parallel algorithm using different options. The Correction option is to fix a discrepancy between the NEMO paper and the NEMO code in the heuristic. The Order option is to order moves such as NEMO or to use the order of the string. Given the results in the table the two options do not seem to matter much.

Table \ref{root} gives the number of problems solved with time using the root parallel GNRPA algorithm with 20 process. The results are slightly worse than when using leaf parallelization with 8 threads. It is due to problems that converge slowly and do not benefit from restarts and where leaf parallelization enables to improve during much longer than root parallelization the best policy. Given enough ressources the best algorithm might be the combination of root and leaf parallelization as in \cite{negrevergne2017distributed}. The parallelization of NRPA proposed by Nagorko \cite{Nagorko19} is also appealing.  

\section{Conclusion}

We experimented with various Monte Carlo Search algorithms for the Inverse Folding problem. We have used very limited domain knowledge, essentially using a small part of the NEMO heuristics for the bias. By applying general Monte Carlo Search heuristics we were able to solve as many problems as NEMO in comparable times.

\section*{Acknowledgment}

Thanks to Fernando Portela for his NEMO program. Tristan Cazenave is supported by the PRAIRIE institute.

\bibliographystyle{splncs04}
\bibliography{main}
\end{document}